\pdfoutput=1

\documentclass[11pt]{article}

\usepackage{authblk}
\usepackage[]{emnlp2021}

\usepackage{times}
\usepackage{latexsym}
\usepackage{kotex}
\usepackage{graphicx}
\usepackage{amsmath}
\usepackage{verbatim}
\usepackage{float}

\usepackage[T1]{fontenc}

\usepackage[utf8]{inputenc}

\usepackage{microtype}

\title{An Evaluation Dataset and Strategy for Building Robust Multi-turn Response Selection Model}

\author[1\thanks{\hspace*{0.6ex} These authors contributed equally to this work.}]{Kijong Han}
\author[2$^{*}$\thanks{\hspace*{0.6ex} This work was done while the author was working at Kakao Enterprise.}]{Seojin Lee}
\author[1]{Wooin Lee}
\author[1]{Joosung Lee}
\author[1\thanks{\hspace*{0.6ex} Corresponding author.}]{Dong-hun Lee}
\affil[1]{Kakao Enterprise, South Korea}
\affil[2]{SK Telecom, South Korea}
\affil[ ]{\texttt {\{mat.h,dan.kes,rung.joo,hubert.std\}@kakaoenterprise.com}}
\affil[ ]{\texttt {seojin.lee@sktair.com}}

\begin{document}
\maketitle
\begin{abstract}
Multi-turn response selection models have recently shown comparable performance to humans in several benchmark datasets. However, in the real environment, these models often have weaknesses, such as making incorrect predictions based heavily on superficial patterns without a comprehensive understanding of the context. For example, these models often give a high score to the wrong response candidate containing several keywords related to the context but using the inconsistent tense. In this study, we analyze the weaknesses of the open-domain Korean Multi-turn response selection models and publish an adversarial dataset to evaluate these weaknesses. We also suggest a strategy to build a robust model in this adversarial environment.
\end{abstract}

\section{Introduction}
\label{section:intro}
Multi-turn response selection is a task that selects the best response among given candidates for a given dialogue context. Response selection models have recently shown comparable performance to humans~\citep{cui2020mutual} in the several in-domain/held-out benchmarks~\citep{lowe2015ubuntu,zhang2018personalizing,dinan2020second}. However, in the actual service environment, these models are often found to have weaknesses. For example, the model gives the highest score to the wrong response, which has high word overlap with the context~\citep{yuan2019multi} or semantically similar to the context~\citep{whang2021response}. 

Held-out evaluation often overestimates the real-world performance of the model~\citep{ribeiro2020beyond}, so adversarial datasets for evaluating weaknesses have been constructed for each task, such as NLI~\citep{naik2018stress,mccoy2019right}, and MRC~\citep{jia2017adversarial, rajpurkar2018know}. A framework for comprehensively evaluating the general linguistic abilities of the model was also studied~\citep{ribeiro2020beyond}.

Several works evaluated adversarial cases for the response selection task~\cite{yuan2019multi, whang2021response}. However, they just automatically generate adversarial responses by copying words in the context. In this study, we analyze the weaknesses of the various aspects of the open-domain Korean Multi-turn Response Selection models and construct an adversarial dataset manually. A total of 2,220 test cases are constructed, and each test case are classified by type.

Neural networks do not generalize well to such an adversarial setting because they tend to use superficial patterns and spurious correlation of the dataset overly, which makes models biased\citep{clark2019don, nam2020learning}. Thus, various debiasing methods have been studied to alleviate this phenomenon~\citep{he2019unlearn, utama2020mind}. In this study, we show that debasing method is also effective in adeversarial evaluation for multi-turn response selection task.

\begin{table*}[t]
\label{table:adv}
{\small
\begin{center}
\begin{tabular}{c|l|l|c}
\hline
\textbf{Type}                                                    & \multicolumn{1}{c|}{\textbf{Context}}                                                                                                                                     & \multicolumn{1}{c|}{\textbf{Adversarial Response}} & \textbf{\# cases} \\ \hline
Repetition                                                       & {[}A{]} I'm hungry / {[}B{]} What do you want to eat?                                                                                                                     & I'm hungry                                         & 400               \\ \hline
\begin{tabular}[c]{@{}c@{}}Negation\\ \end{tabular}    & \begin{tabular}[c]{@{}l@{}}{[}A{]} Wrap up before you go outside / {[}B{]} Why?\\ {[}A{]} It's freezing cold.\end{tabular}                                        & \begin{tabular}[c]{@{}l@{}} Yes, indeed. it's \textbf{not} that cold \\ today. \end{tabular}           & 454               \\ \hline
\begin{tabular}[c]{@{}c@{}}Tense\\ \end{tabular}       & \begin{tabular}[c]{@{}l@{}}{[}A{]} I can't wait to watch "Joker" / {[}B{]} I watched the movie.\\ It was really impressive. / {[}A{]} Wow! I should watch it.\end{tabular} & You really enjoy\textbf{ed} it.                             & 158               \\ \hline
\begin{tabular}[c]{@{}c@{}}Subject-\\ Object\end{tabular}        & \begin{tabular}[c]{@{}l@{}}{[}A{]} I'm in love with BTS / {[}B{]} Why do you like them\\ so much? / {[}A{]} $\phi$(their) Songs are great\end{tabular}                                    & Thanks (for complimenting me)                      & 374               \\ \hline
\begin{tabular}[c]{@{}c@{}}Lexcial \\ Contradiction\end{tabular} & {[}A{]} It's freezing \textbf{cold} today.                                                                                                                                         & \begin{tabular}[c]{@{}l@{}} Yes, indeed. It's way too \textbf{hot} \\ out today. \end{tabular}           & 254               \\ \hline
\begin{tabular}[c]{@{}c@{}}Interrogative\\ Word\end{tabular}   & \begin{tabular}[c]{@{}l@{}}{[}A{]} I saw Jennie today / {[}B{]} What does she look like? /\\ {[}A{]} $\phi$(she) Looks so pretty\end{tabular}                                     & \textbf{Who}'s so pretty?                                   & 236               \\ \hline
Topic                                                          & \begin{tabular}[c]{@{}l@{}}{[}A{]} Isn't the \textbf{weather} nice today? / {[}B{]} Oh, is it?\\ {[}A{]} Yeah, it's sunny and warm.\end{tabular}                                   & Bring your \textbf{umbrella} with you.                      & 344               \\ \hline
\end{tabular}
\end{center}
}
\caption{Examples of adversarial data for each type. $\phi$ denotes a zero anaphora in Korean.}
\end{table*}

In the retrieval-based chatbot system where response selection is used, response candidates are composed as follows. All utterances in the database are used as response candidates~\citep{humeau2019poly}, or part of them filtered through search engines are used~\citep{zhou2020design}. To filter the candidates, machine learning-based embeddings or word-level similarity algorithms(e.g., BM25), which also have weaknesses in an adversarial setting, are used~\citep{zhou2020design}. Therefore, almost every time a response is selected by the actual system, adversarial cases are included in the candidates. Thus, robustness to adversarial cases is more important for response selection task. We also construct a real environment test set and experiment that the model robust to an adversarial case has high performance in the real environment.

\section{Adversarial Test Dataset}
\label{section:dataseet}

We analyze the incorrect responses in the internal service log and categorize the types of frequent errors. There are a total of seven types, and details of each type are as follows.

\textbf{Repetition} An incorrect response repeating one of the utterances in the context.

\textbf{Negation}  A negation is either added to or omitted from a correct response, generating an erroneous response with reversed affirmative or negative meaning. A test set for a negation error intentionally generates a negative response by adding or removing ‘안’ or ‘못’, which are negative adverbs in Korean (short-form negation) or ‘-지 않다,’ ‘-지 못하다, ’ or ‘-지 말다’ which are negative auxiliary predicates in Korean (Long-form negation) in order to test whether the model understands such semantic reversal.

\textbf{Tense} A morpheme or expression marking tense is added to or removed from a correct response, generating an erroneous response in tense that is inconsistent with the given context. A test set for tense errors adds or replaces morphemes or expressions marking the future tense such as ‘-겠-,’  or ones marking the past tense such as‘-었-’ to test whether the model fully understands the context disconnection triggered by such tense change. 

\textbf{Subject-Object}  A test set for subject-object errors generates a response inconsistent with the context due to confusion of the subject and object for a certain action. In particular, since zero anaphora can be found frequently in Korean sentences, incorrect responses are often made because of a failure in identifying the hidden subject of the previous context. This test set uses a subject or an object differently from the ones used in a correct response to examine whether the model fully understands the context disconnection caused by such errors.

\textbf{Lexical Contradiction} A key lexicon of a correct response is replaced with one that holds either conflicting or opposite meaning against the said key lexicon, generating an incorrect response. A test set for lexical contradiction errors replaces a key lexicon in a sentence with an antonym (e.g. hot vs cold) or a word that cannot be used instead (e.g. rain vs snow) to check whether the model understands the precise meaning of such lexicon. 

\textbf{Interrogative Word} A test set for interrogative word errors generates a response in a form of 5W1H questions to ask for information that has already been explicitly or implicitly shared in previous dialogues.

\textbf{Topic} A key sentence or vocabulary is replaced with another sentence or term that does not fit in the previous context even though they frequently appear together in the given topic. While this error is similar to the lexical contradiction error to a certain extent, the replacement words used in this test do not hold conflicting or opposite meanings but instead have less semantic relevance to the context of the previous dialogue (e.g. sunny vs umbrella). The test set assesses whether a model fully understands the fact that while the replacement vocabulary is the one that is frequently used in the same given topic, the response does not correctly reflect the context of the previous dialogue.

Five annotators generate a total of 200 dialogue sessions. For each session $i$, annotators create two correct responses and an arbitrary number($M_i$) of incorrect responses based on the instruction described above. All sessions and responses are reviewed and filtered by experts. We set up one test case to consist of context, one correct response, and one incorrect response. Therefore, $2*M_i$ test cases were extracted for each session, and a total of 2,220 test cases are constructed. It evaluates whether the model gives the correct answer a higher score than the incorrect one for a given context. Statistics and examples are described in Table 1. We release this data set at \url{https://github.com/kakaoenterprise/KorAdvMRSTestData}.
\section{Method}
Suppose that dataset is $D=\left\{(c_i,r_i,y_i)\right\}_{i=1}^{N}$, where $c_i$ denotes a dialogue context, $r_i$ is a response utterance, and $y_i \in \left\{0,1 \right\}$ is a label. The context $c_i=\left\{u_{i,1}, u_{i,2}, ..., u_{i,k_i}\right\}$ consists of sequence of $k_i$ utterances. The label $y_i=1 $ means that $r_i$ is sensible response for context $c_i$.

\subsection{Baseline: Fine-tuning BERT}
\label{section:baseline}
We adopt fine-tuning BERT~\citep{devlin2019bert} as a baseline. In this work, similar to the previous works that fine-tuned BERT for the Multi-turn Response Selection task~\citep{gu2020speaker, whang2020effective, han2021fine}, the input token sequence of BERT $x_i$ is composed as follows.
\begin{equation}\label{eq1}
\begin{aligned}
    x_i = [CLS] u_{i,1} [EOU] ... u_{i,k_{i}} [EOU] [SEP] \\
    r_i [EOU] [SEP]
\end{aligned}
\end{equation}
The [EOU] is a special token indicating that the utterance is over. The final output hidden vector of the [CLS] token in BERT is fed into a fully connected layer with softmax activation. Then, the BERT is fine-tuned to minimize cross entropy loss between the target label and output of this layer.

\subsection{Debiasing Strategy}
\label{section:deb}
In general, correct dialogue response utilizes keywords or topics in the context. Neural networks tend to use such superficial patterns(e.g., keyword, topic) overly, which makes models biased~\citep{clark2019don, nam2020learning}. We see this bias as the main cause of the response selection model's vulnerability to an adversarial environment. Thus, we experimented by applying various debiasing techniques to the response selection task, and DRiFt~\citep{he2019unlearn} was the most effective. The main concept of the debiasing strategy we used is to train a debiased model to fit the residual of the biased model, focusing on examples that cannot be predicted well by biased features only~\citep{he2019unlearn}. Details of the method using DRiFt are as follows.

\begin{figure}[t]\label{figure1}
\label{figure:architecture}
\includegraphics[width=\linewidth]{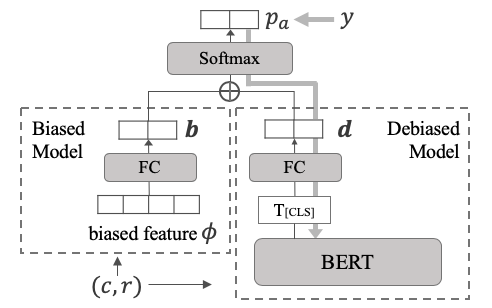}
\caption{Overall architecture for training debiased model utilizing biased model. The grey line represents that gradient is backpropagated only to the debiased model.}
\end{figure}

First, we train an auxiliary biased model using only biased features. The biased model is a single fully connected layer with softmax activation and trained with cross-entropy loss. The biased feature vector used as an input $\phi_i$ is as follows.
\begin{equation}\label{eq2}
\begin{aligned}
\phi_i = [JS_{morph}(c_i,r_i), JS_{morph}(u_{i,k_i},r_i), \\
JS_{wordpiece}(c_i,r_i), JS_{wordpiece}(u_{i,k_i},r_i)]
\end{aligned}
\end{equation}
We use the Jaccard similarity($JS$) between the whole context($c_i$) and response($r_i$) as input features. We also use the JS between the last utterance($u_{i,k_i}$) and $r_i$, because the last utterance is most important~\citep{zhang2018modeling, ma2019triplenet}. We use two tokenizers: the WordPiece~\citep{wu2016google}, and the morpheme analyzer. We assume that these words overlap feature could capture keyword and topic bias.

Second, we train a debiased model utilizing a biased model, as shown in Figure 1. The overall structure of the debiased model is the same as the baseline, but only the learning scheme is different. Let $\textbf{b}$ is output hidden vector of the biased model,  $\textbf{d}$ is output hidden vector of the debiased model, $p_b=softmax(\textbf{b})$, and $p_d=softmax(\textbf{d})$. DRiFt method minimize cross entropy loss between $p_a=softmax(\textbf{b}+\textbf{d})$ and target labels. Thus, the loss function is defined as follows.

\begin{table*}[t]
\begin{minipage}{.52\linewidth}
    {\small
    \begin{tabular}{c|c|c|ccc}
    \hline
    \multicolumn{1}{l|}{} & Train & Val & \begin{tabular}[c]{@{}c@{}}In-domain\\ Test\end{tabular} & \begin{tabular}[c]{@{}c@{}}Adv\\ Test\end{tabular} & \begin{tabular}[c]{@{}c@{}}Real\\ Test\end{tabular} \\ \hline
    \# pairs              & 500K  & 10K & 10K                                                      & 4,440                                              & 5,490                                               \\
    \# cands              & 2     & 10  & 10                                                       & 2                                                  & 10                                                  \\
    pos:neg               & 1:1   & 1:9 & 1:9                                                      & 1:1                                                & 4.4:5.6                                             \\
    \# turns              & 4.6   & 4.5 & 4.6                                                      & 2.9                                                & 3.3                                                 \\ \hline
    \end{tabular}
    \label{table:dataset}
    \caption{Statistics for each dataset.}
    }
\end{minipage}
\begin{minipage}{.48\linewidth}
     {\small
    \begin{tabular}{r|ccc}
    \hline
    Method & In-domain & Adversarial & Real Env. \\ \hline
    baseline              & 86.4±0.5  & 39.4±1.7    & 36.2±0.9     \\
    +deb           & 85.4±0.7  & 43.5±2.3    & 36.5±0.6     \\
    +UMS                  & \textbf{87.5±0.7}  & 42.9±2.5    & 38.8±0.6     \\
    +UMS+deb      & 87.0±0.7  & \textbf{47.2±2.1}    & \textbf{40.4±0.7}    \\
    \hline
    \end{tabular}
    \label{table:peformance}
    \caption{Overall performance of each method.}
    }
\end{minipage}
\end{table*}

\begin{table*}[!hpt]
{\small
\begin{center}
\begin{tabular}{c|ccccccc}
\hline
Type     & Repetition & Negation & Tense    & \begin{tabular}[c]{@{}c@{}}Subject\\ -Object\end{tabular} & \begin{tabular}[c]{@{}c@{}}Lexical \\ Contradiction\end{tabular} & \begin{tabular}[c]{@{}c@{}}Interrogative \\ Word\end{tabular} & Topic    \\ \hline
baseline & 12.9±2.5   & 36.1±2.3 & 41.8±2.3 & 55.0±1.9                                                  & 41.1±1.8                                                         & 46.1±2.2                                                      & 50.7±1.8 \\
+deb     & 26.2±5.1   & \textbf{40.6±2.8} & 43.0±2.6 & 56.2±1.9                                                  & \textbf{43.9±1.6}                                                         & 45.5±3.2                                                      & 51.8±1.9 \\
+UMS     & 26.2±6.2   & 34.5±1.8 & 45.3±2.0 & 61.2±2.9                                                  & 41.6±3.1                                                         & \textbf{46.4±2.0}                                                      & 51.1±1.6 \\
+UMS+deb & \textbf{40.0±5.4}   & 38.5±1.1 & \textbf{46.9±2.3} & \textbf{63.9±3.0}                                                  & 43.3±1.7                                                         & 45.9±2.6                                                      & \textbf{52.8±1.7} \\ \hline
\end{tabular}
\end{center}
}
\caption{Performance for each adversarial type.}

\end{table*}

\begin{equation}\label{eq3}
\begin{aligned}
Loss = -\log{p_a(y_i)} = \\ -\log{p_b(y_i)} -\log{p_d(y_i)}+\log{\sum_{l=0}^{L-1} {p_b(l)p_d(l)}}
\end{aligned}
\end{equation}
$L$ is the number of classification classes(2 for this task). The gradient is backpropagated only to the debiased model. The last term encourages output from the debiased model $p_d$, to have minimal projection on output from the biased model $p_b$~\citep{he2019unlearn}. Derivation of equation \ref{eq3} is in Appendix A. At test time, only debiased model is used.

\subsection{Combination with Multi-task Learning}
\label{section:UMS}
Recently, self-supervised learning approaches have shown state-of-the-art performance in the response selection task~\citep{whang2021response,xu2021learning}. These works devise auxiliary tasks to understand the dialogue better and train the model in a multi-task manner. The final loss function in these methods is the weighted sum of losses of auxiliary tasks and main task (i.e., determine given response is a sensible response to the context). Thus, debiasing strategy could be easily combined with these methods by replacing the loss function of the main task with equation \ref{eq3}. We also experiment with self-supervised learning approach UMS~\citep{whang2021response}, and we show that it is also effective in not only in-domain but also adversarial and real environments.

\section{Experiments and Results}

\subsection{Experiment Setup}
We construct an experimental dataset using the corpus that we produced in-house and the public Korean dialogue corpus\footnote{\url{https://corpus.korean.go.kr}}. We split these corpora into three, and each is for training, validation, and test. Statistics of each dataset are described in Table 2. \#pairs denote the number of context-response pairs, \#cands denotes the number of candidates per context, pos:neg denotes the ratio of positive and negative responses in candidates, and \#turns denote the average turns per context. Details on the construction are as follows.

\textbf{Train, valid, and in-domain test} The last utterance of the dialogue session is used as a positive response and the rest as context. Negative responses are randomly chosen from the other dialogue. 

\textbf{Adversarial test} It is described in the Section~\ref{section:dataseet}.

\textbf{Real environment test} In a real environment, response candidates are not sampled randomly but are sampled through a search system~\citep{zhou2020design}, or all utterances without sampling are used as candidates~\citep{humeau2019poly}. There are many adversarial negatives in this situation, as described in Section~\ref{section:intro}. We build a dataset by simulating this situation in a similar way to the previous works~\citep{wu2017sequential, zhang2018modeling}. 

We take a dialogue session from the test corpus and internal service log as context. We trained a bi-encoder-based context and response embedding model~\citep{humeau2019poly} and indexed embeddings of all utterances in the corpus. Then, we retrieve the top 10 utterances based on the similarity score between context embedding as response candidates. For each response, three annotators labeled whether it is sensible to the context. The response determined by more than two people as sensible was selected as the positive response.

\subsection{Results}
We measure the performance ten times for each model and report the mean and standard deviation in Table 3. See Appendix B for details of training. The baseline is a fine-tuned BERT described in Section~\ref{section:baseline}. "deb" denotes a debiasing strategy described in Section~\ref{section:deb}. UMS denotes a self-supervised multi-task learning method described in Section~\ref{section:UMS}. Precision@1 is used as an evaluation metric for all test sets.

Debiasing strategy significantly improves adversarial test performance in both baseline and UMS model; it achieves absolute improvements of 4.1\% and 4.3\% on baseline and UMS. A decline in performance is observed in the in-domain test; -1.0\% and -0.5\% on baseline and UMS, as the DRiFt debiasing method~\citep{he2019unlearn} shows a slight performance degradation in the in-domain test. However, It improves performance in the comprehensive real environment test; +0.3\% and +1.6\% on baseline and UMS. This supports our argument that robustness to adversarial cases is important in the response selection task. Additionally, +UMS+deb outperforms +deb in all test set. From this, it can be seen that the debiasing strategy and UMS have a synergistic effect.

The performance of each adversarial type is reported in Table 4. Since we used word-level Jacard Similarity as a biased feature, the debiasing strategy shows huge performance improvement in the Repetition type, which simply uses word sequence in context as a negative response. There is no improvement in the Interrogative Word type. We assume that the reason for it is that this type is difficult because it requires understanding all 5W1H from the context.

\section{Conclusion}
We analyze the weaknesses of the open-domain Korean Multi-turn Response Selection models and publish an adversarial dataset to evaluate these weaknesses. We suggest a strategy to build a robust model to an adversarial and real environment with the experimental results. We expect that this work and dataset will help improve the response selection model.

\section{Ethical Considerations}
The adversarial dataset we publish is generated manually. All sessions and responses in the dataset are reviewed and filtered by the experts, and we also considered ethical issues in this process. Thus, there is no hate speech or privacy issue in our dataset.

\bibliography{emnlp2021}
\bibliographystyle{acl_natbib}

\appendix
\clearpage

\section{Derivation of Loss Function}
\label{sec:appendix_a}
Let $b$ is output hidden vector of the biased model, $d$ is output hidden vector of the debiased model, $y_i \in \left\{0,1 \right\}$ is the label value, $p_b=softmax(b)$, $p_d=softmax(d)$, and $p_a=softmax(b+d)$.

\begin{align*}
Loss = -\log p_a(y) \\
= -\log e^{b_y+d_y} + \log \sum_{l} e^{b_l+d_l}\\
= -\log e^{b_y} - \log e^{d_y} + \log \sum_{l} e^{b_l}e^{d_l} \\
= -\log e^{b_y} - \log e^{d_y} + \log \sum_{l} e^{b_l}e^{d_l} \\
+ \log \sum_{l} e^{b_l} - \log \sum_{l} e^{b_l} \\
+ \log \sum_{l} e^{d_l} - \log \sum_{l} e^{d_l} \\
= -(\log e^{b_y} - \log \sum_{l} e^{b_l}) \\
-(\log e^{d_y} - \log \sum_{l} e^{d_l}) \\
+ (\log \sum_{l} e^{b_l}e^{d_l} - \log \sum_{l} e^{b_l} \sum_{l} e^{d_l}) \\
= -\log\frac{e^{b_y}}{\sum_{l} e^{b_l}} - \log\frac{e^{d_y}}{\sum_{l} e^{d_l}} \\
+ \log \sum_{l} \frac{e^{b_l}e^{d_l}}{\sum_{l} e^{b_l} \sum_{l} e^{d_l}} \\
= -\log p_b(y) - \log p_d(y) + log \sum_{l} p_b(l)p_d(l)  \\
\end{align*}

\section{Training Details}
\label{sec:appendix_b}
The biased model, which consists of a single fully connected layer, is trained using the AdamW optimizer with a learning rate of 5e-4 and for 3 epochs. BERT-based models, including baseline, UMS, and debiased models, are trained using the AdamW optimizer with a learning rate of 2.5e-5 and for 3 epochs on 4 Nvidia Volta v100 GPU. The batch size is 128 for every model. We train and evaluate 10 times for each model and calculate mean and standard deviation. For each model, a checkpoint that shows the best performance in the real environment is selected for performance measure. 

\end{document}